# On Local Optima in Learning Bayesian Networks

Jens D. Nielsen, Tomáš Kočka and Jose M. Peña
Department of Computer Science
Aalborg University, Denmark
{dalgaard, kocka, jmp}@cs.auc.dk

## Abstract

This paper proposes and evaluates the $k$-greedy equivalence search algorithm (KES) for learning Bayesian networks (BNs) from complete data. The main characteristic of KES is that it allows a trade-off between greediness and randomness, thus exploring different good local optima when run repeatedly. When greediness is set at maximum, KES corresponds to the greedy equivalence search algorithm (GES). When greediness is kept at minimum, we prove that under mild conditions KES asymptotically returns any inclusion optimal BN with nonzero probability. Experimental results for both synthetic and real data are reported showing that KES finds a better local optimum than GES considerably often. Additionally, these results illustrate that the number of different local optima is usually huge.

## 1 INTRODUCTION

Learning Bayesian networks from data has been widely studied for the last few years. Two approaches to learning have been developed: one tests conditional independence constraints while the other searches the space of models using a score. In both of these frameworks asymptotically optimal learning algorithms have been developed under the faithfulness assumption, the PC algorithm (Spirtes et al. 1993) for the constraint based approach and the greedy equivalence search algorithm (Chickering 2002) for the space search approach. In this paper we deal with the space search approach, usually called *model selection*.

A model selection procedure usually consists of three components: (1) a *neighborhood*, (2) a *scoring criterion* and (3) a *search strategy*. The neighborhood of a model restricts the search to a small part of the search space around that model, usually defined by means of local transformations of the model representative.[1] The scoring criterion evaluates the quality of a model and it is usually required to be score equivalent, locally consistent and decomposable. The search strategy selects a new model, from those in the neighborhood of the current best model, based on the scoring criterion.

Chickering (2002) shows that the greedy equivalence search algorithm (GES), using the inclusion boundary neighborhood and a locally consistent scoring criterion, is asymptotically optimal under the faithfulness assumption. Several of these assumptions are not strictly necessary to obtain the optimality: the neighborhood used can be slightly reduced, the faithfulness assumption is replaced by a weaker assumption, namely the composition property assumption, in (Chickering and Meek 2002) yielding inclusion optimal GES and, finally, the greediness of GES is not needed.

The same optimality result as for GES holds for any BN learning algorithm that, using the inclusion boundary neighborhood and a locally consistent scoring criterion, strictly increases the score in every step and stops when there is no model with a higher score in the neighborhood of the current best model. In other words, GES chooses in every step the best scoring neighbor of the current best model, while choosing any of its neighbors that are better than the current best model is sufficient for asymptotic optimality. Thus, the most we can relax the greediness of GES in favor of randomness while keeping the optimality is what we call the stochastic equivalence search algorithm (SES). SES randomly selects in every step a model, from those in the inclusion boundary neighborhood of the current best model, with higher score than the current best

---

[1] We completely leave out the question of the representation of the model. It can be an acyclic directed graph, an essential graph, a pattern or even a largest chain graph. The representation is not crucial, even though some representations are more efficient than others for generating some neighborhoods.

model. It stops when there is no neighbor better than the current best model.

A natural generalization of both SES and GES consists in making the degree of greediness a parameter of the learning algorithm. We call this algorithm the $k$-greedy equivalence search algorithm (KES), where the parameter $k$ specifies the degree of greediness. SES and GES are, therefore, special cases of KES. KES combines the greedy bias towards promising solutions of GES with the randomness of SES, which enables it to visit different local optima when run repeatedly. Consequently, KES can outperform GES in practice, as it is known that the behavior of the latter may be suboptimal if several locally optimal models exist (Chickering and Meek 2002).

In the next section we introduce basic concepts of probability distributions, Bayesian network models and their learning. In Section 3 we define KES and prove some theoretical results. Section 4 describes our implementation of KES, the data we use in the experiments and the results obtained. We conclude in Section 5.

## 2 BASIC CONCEPTS

Throughout the paper the symbol $V$ denotes a nonempty finite set of discrete variables that are identified with nodes of graphs. A *directed graph* $G$ over a set of nodes $V$ is specified by a collection of *arcs* $A_G$, i.e. a collection of ordered pairs of distinct nodes in $V$. A directed graph $G$ is an *acyclic directed graph* (DAG) if there is no directed cycle formed by the arcs in $G$. If there is an arc $X \to Y$ in $G$, then the node $X$ is a *parent* of the node $Y$. The set of parents of $Y$ in $G$ is denoted by $Pa_G(Y)$ and when the graph is clear from context by $Pa(Y)$ only. The union of a node $X$ and its parents is called a *family* and is denoted $Fa(X) = \{X\} \cup Pa(X)$. An arc $X \to Y$ is *covered* in a DAG $G$ if $Pa(Y) = Fa(X)$ in $G$. For every DAG $G$ there exists a *causal order* $CO(G)$ of the nodes $V$ such that $X$ precedes $Y$ in $CO(G)$ if $X \to Y$ in $G$. We denote by $Pre(X)$ the set of nodes that precede the node $X$ in a causal order.

A *Bayesian network* (BN) is a pair $(G, \theta)$, where $G$ is a DAG and $\theta$ are parameters specifying the conditional probability distribution for each node $X \in V$ given its parents, $p(X|Pa(X))$. A BN represents a joint probability distribution $p(V)$ through the factorization $p(V) = \prod_{X \in V} p(X|Pa(X))$. A BN *model* $M(G)$ is the set of all the joint probability distributions $p(V)$ that can be represented by a parameterization $\theta$ of the BN $(G, \theta)$.

Model $M_1$ *includes* model $M_2$, denoted $M_2 \subseteq M_1$, if for every parameterization $\theta_2$ of $M_2$ there exists a parameterization $\theta_1$ of $M_1$ that represents the same joint probability distribution over the (observed) variables $V$. Two models $M_1$ and $M_2$ are the same if $M_1$ includes $M_2$ and $M_2$ includes $M_1$, i.e. $M_1 = M_2$. $M_1$ *strictly includes* $M_2$, denoted $M_2 \subset M_1$, if $M_2 \subseteq M_1$ and $M_1 \neq M_2$. Two DAGs $G_1$ and $G_2$ are *equivalent* if they represent the same model, i.e. $M(G_1) = M(G_2)$. It is shown by Chickering (1995) that two DAGs $G_1$ and $G_2$ are equivalent iff there is a sequence of covered arc reversals that converts $G_1$ into $G_2$. A model $M$ is *inclusion optimal* w.r.t. a joint probability distribution $p$ if $M$ includes $p$ and there is no model strictly included in $M$ that includes $p$.

All the joint probability distributions in a BN model $M(G)$ satisfy certain conditional independence constraints among the variables that can be read from the DAG $G$ by means of d-separation, one says that they are *globally Markov* w.r.t. $G$. See Lauritzen (1996) for the definition of d-separation. Joint probability distributions that do not satisfy any other conditional independence than those enforced by d-separation in $G$ are called *faithful* to $G$. Note that among others the conditional independencies $X \perp\!\!\!\perp Pre(X) \backslash Pa(X) | Pa(X)$ for all the nodes $X \in V$ are enforced by d-separation in $G$ for any causal order $CO(G)$ in $G$. A joint probability distribution $p$ that satisfies these conditional independencies is said to satisfy the *local ordered Markov property* w.r.t. $CO(G)$ (Cowell et al. 1999), and it is included in the model $M(G)$, i.e. $p \in M(G)$.

Conditional independencies holding in a joint probability distribution satisfy semi-graphoid axioms: *triviality* $X \perp\!\!\!\perp \emptyset | Z$, *symmetry* $X \perp\!\!\!\perp Y | Z \Rightarrow Y \perp\!\!\!\perp X | Z$, *decomposition* $X \perp\!\!\!\perp YU | Z \Rightarrow X \perp\!\!\!\perp Y | Z$, *weak union* $X \perp\!\!\!\perp YU | Z \Rightarrow X \perp\!\!\!\perp Y | ZU$ and *contraction* $X \perp\!\!\!\perp Y | ZU \land X \perp\!\!\!\perp U | Z \Rightarrow X \perp\!\!\!\perp YU | Z$. Note that from contraction, weak union and decomposition follows the so called *block independence lemma* $X \perp\!\!\!\perp Y | ZU \land X \perp\!\!\!\perp U | Z \Leftrightarrow X \perp\!\!\!\perp YU | Z$. In addition, any joint probability distribution faithful to some DAG (as well as many more distributions) satisfies the *composition* property $X \perp\!\!\!\perp Y | Z \land X \perp\!\!\!\perp U | Z \Rightarrow X \perp\!\!\!\perp YU | Z$.

One uses data to select among different models according to some scoring criterion that assigns a score $S(M)$ to a model $M$. Sometimes, it is convenient to assign a score $S(G) = S(M(G))$ to a representative DAG of the model, too. If a scoring criterion assigns the same value to all the DAGs representing the same model, then we say that the scoring criterion is *score equivalent*. A scoring criterion is *locally consistent* if the score assigned to a DAG $G$ for some data i.i.d. sampled from a joint probability distribution $p$ asymptotically always increases by removing an arc in $G$, unless this arc removal adds a conditional independence constraint to the model that does not hold in $p$. A scoring criterion is *decomposable* if it can be expressed as a sum over all the nodes $X \in V$ of some function of their families $Fa(X)$. We consider in this paper the two most common scoring criteria, namely the Bayesian information criterion (BIC) and the Bayesian marginal likelihood with "uniform" Dirichlet prior (BDeu). Both are score equivalent, locally consistent and decomposable.

The *inclusion boundary* $IB(M_1)$ of a model $M_1$ is the union of the lower and upper inclusion boundaries, $LIB(M_1)$ and $UIB(M_1)$, respectively. $LIB(M_1)$ is the set of models $M_2$ that are strictly included in $M_1$ and such that no model strictly included in $M_1$ strictly includes $M_2$. Likewise, $UIB(M_1)$ is the set of models $M_2$ that strictly include $M_1$ and such that no model strictly including $M_1$ is strictly included in $M_2$. $IB(M(G))$ is characterized by Chickering (2002) to be the set of models represented by all those DAGs that can be obtained by adding or removing a single arc from any DAG $G^*$ equivalent to $G$. A more complex characterization of the inclusion boundary of a BN model, which enables an efficient generation of the inclusion boundary, can be done using essential graphs.

**Theorem 1** *(Chickering 2002, Lemmas 9 and 10) The greedy equivalence search algorithm (GES) using inclusion boundary neighborhood, locally consistent scoring criterion and fully observed learning data i.i.d. sampled from a joint probability distribution faithful to a DAG $G$ asymptotically always discovers the model $M(G)$.*

The version of GES we use in this paper considers in each step the whole inclusion boundary of the current best model and is described as a variant of the original GES in (Chickering 2002). The original version of GES (Meek 1997) is a two-phase algorithm using first only the upper inclusion boundary and afterwards only the lower inclusion boundary. The theorem above holds for both versions of GES. Moreover the two algorithms usually proceed in exactly the same way in practice. Thus we use the conceptually simpler alternative, despite the original GES is closer to the optimality proof and performs slightly less operations. Note that even the original GES does unnecessarily many operations in the first phase. One does not have to use the whole upper inclusion boundary, it is enough to be able to add a single arc for any pair of non-adjacent nodes.

As the faithfulness assumption is not realistic, Chickering and Meek (2002) propose replacing it by a weaker assumption, the composition property assumption.

**Theorem 2** *(Chickering and Meek 2002, Theorem 4) The greedy equivalence search algorithm (GES) using inclusion boundary neighborhood, locally consistent scoring criterion and fully observed learning data i.i.d. sampled from a joint probability distribution $p$ satisfying the composition property asymptotically always discovers a model that is inclusion optimal w.r.t. $p$.*

Chickering and Meek (2002) provide an example of an undirected graphical model for which joint probability distributions faithful to it satisfy the composition property and exhibit two inclusion optimal BN models. We extend it by a particular parameterization for which GES is not asymptotically optimal, it asymptotically returns the locally but not globally optimum.

**Example 1** *Let $X$ be a variable with four states and $Y$, $Z$ and $U$ be three binary variables. Let $p$ be a joint probability distribution over $XYZU$, satisfying the conditional independencies $X \perp Z|YU$ and $Y \perp U|XZ$ and having marginal joint probability distributions $p(XY) = (0.22, 0.03, 0.22, 0.03; 0.03, 0.22, 0.03, 0.22)$, $p(XU) = (0.22, 0.22, 0.03, 0.03; 0.03, 0.03, 0.22, 0.22)$ and $p(YZ) = p(UZ) = (0.35, 0.15; 0.15, 0.35)$. There are two inclusion optimal BN models w.r.t. $p$, $M(G_1)$ and $M(G_2)$ with $A_{G_1} = \{X \to Y, X \to U, Y \to Z, U \to Z, Y \to U\}$ and $A_{G_2} = \{X \to Y, X \to U, Y \to Z, Z \to U, X \to Z\}$. The two models have different number of parameters, 19 for $M(G_1)$ and 23 for $M(G_2)$. Thus, $M(G_1)$ is globally optimal while $M(G_2)$ is not.*

## 3 KES ALGORITHM

As seen above, GES may be asymptotically suboptimal when several inclusion optimal models exist. A straightforward solution to this problem consists in the addition of randomness to the BN learning algorithm. In this section we introduce the $k$-greedy equivalence search algorithm (KES), which allows the user to trade off between greediness and randomness. KES is formally described as follows:

```
KES (k∈[0,1])
M = empty graph model
repeat
    B = set of models in IB(M) with
        higher score than the model M
    if |B| > 0 then
        C = random subset of the set B
            with size max(1,|B|·k)
        M = the highest scoring model
            from the set C
    else return(M)
```

It is easy to see that KES includes SES and GES as special cases with settings $k = 0$ and $k = 1$, respectively. We will show that the nice properties of GES described in Theorems 1 and 2 hold for KES, too.

**Theorem 3** *The k-greedy equivalence search algorithm (KES) using inclusion boundary neighborhood, locally consistent scoring criterion and fully observed learning data i.i.d. sampled from a joint probability distribution faithful to a DAG $G$ asymptotically always discovers the model $M(G)$.*

**Proof:** The high level idea behind the proof of Theorem 1, see (Chickering 2002, Lemmas 9 and 10), is that asymptotically, for data faithful to a DAG $G$ and locally consistent score, for any model $M$ different from $M(G)$ there is another model in $IB(M)$ with higher score. This is explicitly shown in (Castelo and Kočka 2003, Theorem 3.4). Thus, any algorithm that increases the score in each step and stops only when there is no model in $IB(M)$ with higher score than the current best model $M$, always finds the model $M(G)$. This holds because there is a finite number of different models, and increasing the score in each step prevents the algorithm from visiting the same model twice. It is obvious that this proof applies to KES equally well as to GES, for which it was first derived. □

**Theorem 4** *The k-greedy equivalence search algorithm (KES) using inclusion boundary neighborhood, locally consistent scoring criterion and fully observed learning data i.i.d. sampled from a joint probability distribution $p$ satisfying the composition property asymptotically always discovers a model that is inclusion optimal w.r.t. $p$.*

**Proof:** The high level idea behind the proof of Theorem 2, see (Chickering and Meek 2002, Theorem 4), is that asymptotically, for data sampled from a joint probability distribution $p$ satisfying the composition property and locally consistent score, for any model $M$ that is not inclusion optimal w.r.t. $p$ there is another model in $IB(M)$ with higher score. Therefore, the result applies to KES, too. □

Moreover, KES with $k = 0$, i.e. SES, can asymptotically find any inclusion optimal model. In practice, SES examines all the local optima if run repeatedly enough times.

**Theorem 5** *The stochastic equivalence search algorithm (SES) using inclusion boundary neighborhood, locally consistent scoring criterion and fully observed learning data i.i.d. sampled from a joint probability distribution $p$ satisfying the composition property asymptotically discovers with nonzero probability any model that is inclusion optimal w.r.t. $p$.*

**Proof:**[2] Let $M(G)$ be any inclusion optimal model w.r.t. $p$. Let us denote by $e = |A_G|$ the cardinal-ity of $A_G$, i.e. the number of arcs in $G$. We will prove the theorem by constructing a sequence of models $M(G_0), \ldots, M(G_e)$ where $G_0$ is the empty graph, $G_e = G$ and each $G_i$ is obtained from $G_{i-1}$ by adding an arc that increases the score. Obviously $M(G_i)$ is in $IB(M(G_{i-1}))$ and, thus, considered by SES with nonzero probability. We denote by $A_{G_i}$ the arcs in $G_i$. It holds for each $G_i$ in the sequence that $A_{G_i} \subseteq A_G$.

---

[2]This proof proceeds similarly to the proof of Theorem 2, see (Chickering and Meek 2002, Theorem 4). However,

It is easy to see that $G_0$ satisfies the condition $A_{G_0} \subseteq A_G$. Thus, we will use the induction hypothesis $A_{G_i} \subseteq A_G$, and we will show that there is a DAG $G_{i+1}$ obtained from $G_i$ by adding an arc for which $A_{G_{i+1}} \subseteq A_G$ and $S(G_{i+1}) > S(G_i)$ for all $i < e$.

It follows from $i < e$ and $A_{G_i} \subseteq A_G$ that $M(G_i) \subset M(G)$. Because the model $M(G)$ is inclusion optimal, $p \in M(G)$ and $p \notin M(G_i)$. It follows from $p \notin M(G_i)$ that there is a node $X$ for which the conditional independence $X \perp\!\!\!\perp Pre(X) \backslash Pa_{G_i}(X) | Pa_{G_i}(X)$ is enforced by d-separation in $G_i$ but it does not hold in $p$. Moreover, $X \perp\!\!\!\perp Pre(X) \backslash Pa_G(X) | Pa_G(X)$ is enforced by d-separation in $G$ and, thus, it holds in $p$, too. Note that $Pre(X)$ are the same in both $G$ and $G_i$ if any causal order $CO(G)$ in $G$ is used as a causal order in $G_i$ as well. This is possible because $A_{G_i} \subseteq A_G$. Then, $X \not\perp\!\!\!\perp Pre(X) \backslash Pa_{G_i}(X) | Pa_{G_i}(X)$ in $p$ together with $X \perp\!\!\!\perp Pre(X) \backslash Pa_G(X) | Pa_G(X)$ in $p$ yields $X \not\perp\!\!\!\perp Pa_G(X) \backslash Pa_{G_i}(X) | Pa_{G_i}(X)$ in $p$ by the block independence lemma. It follows, by composition, that there exists a node $Y \in Pa_G(X) \backslash Pa_{G_i}(X)$ for which $X \not\perp\!\!\!\perp Y | Pa_{G_i}(X)$ in $p$. Thus, we can add the arc $Y \to X$ to $G_i$ and obtain a DAG $G_{i+1}$ that has asymptotically higher score (because we have removed a conditional independence that does not hold in $p$ and we use a locally consistent score). Moreover, $A_{G_{i+1}} \subseteq A_G$ (note that $Y \in Pa_G(X)$).

We have proven above that the algorithm can asymptotically get to any inclusion optimal model $M(G)$. It follows from the definitions of inclusion optimality and locally consistent score that the algorithm asymptotically stops in the model $M(G)$ because all the models in $IB(M(G))$ have lower score. □

The bad news, which limits the practical applicability of SES, is that the number of inclusion optimal models for a domain with $n$ variables can be exponential in $n$.

**Example 2** *The undirected model over four variables in Example 1 encodes a joint probability distribution having two inclusion optimal models. Consider a model $M$ that consists of $n$ such disjoint undirected models. $M$ encodes a joint probability distribution where all these $n$ groups of four variables are indepen-*

---

note that Chickering and Meek use the local Markov property for their proof, while we need to resort to the local ordered Markov property in order to prove our theorem.

*dent and, thus, each inclusion optimal model consists of $n$ disjoint submodels. For each of these $n$ groups of four variables there are two possible inclusion optimal submodels and any combination of these is an inclusion optimal model. Then, there are $2^n$ inclusion optimal models for a domain of $4 \cdot n$ variables.*

## 4 EVALUATION

In this section we empirically evaluate the effectiveness of SES and KES ($k \neq 0, 1$) with respect to GES, for both synthetic and real data. We start by describing the implementation of KES. Then, the databases involved in the evaluation are introduced. Finally, we report and discuss the results obtained.

### 4.1 IMPLEMENTATION

So far we have studied the theoretical properties of KES (including SES), regardless of the representation of the models in the search space. However, we need to adopt a certain model representation scheme in order to evaluate its effectiveness in practice. As mentioned before, common representations include DAGs, essential graphs, patterns and largest chain graphs. For the sake of simplicity, we represent each model in the search space by a DAG. This approach involves a major difficulty: there is no efficient way of generating $IB(M(G))$ in each iteration of KES using DAGs, where the DAG $G$ represents the current best model.

Recall from Section 3 that KES first generates the set $B$ of models in $IB(M(G))$ that are better than $M(G)$ and, then, it samples a random subset $C$ of $B$ of a defined size. As we cannot produce $IB(M(G))$ efficiently in our implementation of KES, we first draw a random sample of a defined size from $IB(M(G))$ and, then, we select the models than are better than $M(G)$ out of those in the sample, i.e. we swap in practice the first two steps of KES. This way of working has an impact on the implementation of the stopping criterion of KES: instead of halting when no model in $IB(M(G))$ has higher score than $M(G)$, KES stops in practice when $M(G)$ has not been improved for a sufficiently large number of consecutive iterations.

Fortunately, there is a simple way of generating at random any member of $IB(M(G))$ with nonzero probability (Castelo and Kočka 2003): choose at random a covered arc in $G$ and reverse it, repeat this covered arc reversal operation (CAR) several times and, then, choose at random a pair of nodes and either remove the arc between them or add an arc between them without creating a directed cycle. Specifically, we use a descendant matrix to check for directed cycles. If we need to generate more than one member of $IB(M(G))$, then we can repeat the process above. Note that we do not have to do the repeated CAR each time a member of $IB(M(G))$ is to be produced. Instead, we may perform CARs less frequently and still randomly generate a member of $IB(M(G))$ as far as the total number of CARs is high enough.

It should be mentioned that the implementation of KES described above may sample from $IB(M(G))$ the same model more than once per iteration. To neutralize this risk and draw the specified amount of distinct models in each step of KES, the parameter $k$ is internally translated into $k^*$ in the implementation of KES, so that the mean percentage of distinct models sampled from $IB(M(G))$ by generating at random $k^* \cdot |IB(M(G))|$ models with the possibility of repeating equals $k$. For example, $k = 0.4$ corresponds to $k^* = 0.5$. Note that in order to guarantee that KES with $k = 1$ behaves as GES in practice, i.e. the best scoring model in $IB(M(G))$ is selected in each iteration, $k^*$ must equal infinity which is, of course, impractical. We checked in all our experiments that we used a sufficiently high value for $k^*$ when KES simulated GES ($k^* = 20$ was enough). As the exact value of $|IB(M(G))|$ is difficult to compute, we approximate it by the number of arcs that can be added to an empty graph over $V$, i.e. $|V| \cdot (|V| - 1)$. This is a good approximation as shown in (Chickering 2002) and thanks to the fact that connectivity components in essential graphs representing common models tend to be small.

We are aware that using essential graphs to represent the models in the search space may avoid many of the implementation issues discussed above and, consequently, result in a more efficient implementation of KES. However, this approach is discarded due to its higher complexity. Moreover, we are mainly concerned in this paper with the effectiveness of KES.

Finally, it is a good idea in any implementation of a BN learning algorithm to store computed scores in a cache with the purpose of avoiding computing them more than once. This is particularly necessary in our implementation of KES: only by using some cache mechanism we can afford generating the same models as many times as it may occur when KES simulates GES in our implementation, because the score is computed from data only the first time while all the repeated computations are done extremely fast by accessing the cache. Specifically, our implementation of KES uses a tree-like structured cache to store the scores computed for any family, i.e. we take advantage of the decomposability of the BIC and the BDeu. Each entry $E$ of the cache tree stores the score for the node specified in the first level of the branch from the root entry to $E$ given the parents specified by the remaining entries in the branch. It should be mentioned that we consider an arbitrary ordering of the nodes in $V$ according to which the parents are always ordered to access the cache. This relates each family to a unique entry of the cache tree, and vice versa. The structure of the cache described here is similar to an R-tree index, but it is fixed and reasonably balanced. Only those parts of the cache tree that are needed are allocated in memory. This cache provides an extremely fast alternative to computing the score from data.

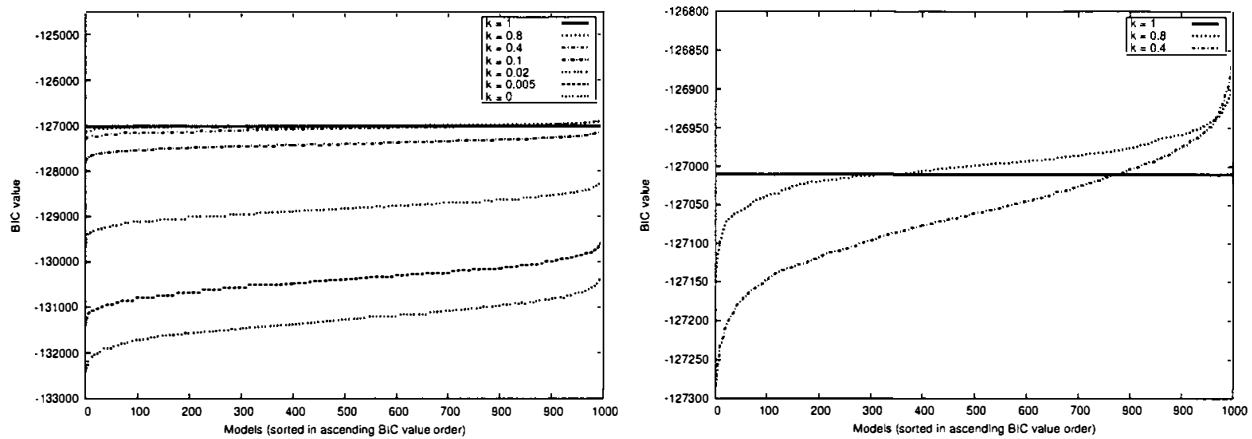

Figure 1: Effectiveness of KES for the ALL database when $k = 0, 0.005, 0.02, 0.1, 0.4, 0.8, 1$ (left), and a close-up thereof for $k = 0.4, 0.8, 1$ (right).

### 4.2 DATABASES

The first synthetic database in the evaluation is the widely studied Alarm database (Herskovits 1991). It contains 20000 cases, sampled from a BN representing potential anesthesia problems in the operating room. The generative model has 37 nodes and 46 arcs, and each node has from two to four states. As reported by Cooper and Herskovits (1992), one of the arcs in the generative model is not supported by the data.

The second synthetic database in the evaluation is named the Trap database and consists of 20000 cases, sampled from the joint probability distribution in Example 2 with $n = 10$, i.e. 40 nodes in the model.

The first two real databases in the evaluation are obtained by preprocessing the Leukemia database (Golub et al. 1999), 72 samples from leukemia patients with each sample being characterized by the expression levels of 7129 genes, as follows. First, gene expression levels are discretized into three states via an information theory based method (Beibel 2000). Then, the discretized database is split into two auxiliary databases: one containing the data of the 47 patients suffering from acute lymphoblastic leukemia (ALL), and the other containing the data of the 25 patients suffer-ing from acute myeloid leukemia (AML). Finally, these two databases are transposed, so that the 7129 genes are the cases and the measurements for the corresponding patients are the attributes. The resulting databases are simply denoted ALL and AML, respectively, in the forthcoming discussion. It should be mentioned that the cases in ALL and AML are treated as i.i.d., although some genes may be co-regulated and, thus, some cases may be correlated. This simplifies the analysis and may not change the essence of the results. In fact, this approach is commonly taken in gene expression data analysis (e.g. Ben-Dor et al. (2001)).

The third real database in the evaluation comes from the CoIL Challenge 2000 (Putten and Someren 2000), and contains information on customers of an insurance company. The database in the evaluation corresponds to the training data in the challenge, but restricted to attributes 2, 43 and 65-86, i.e. 5822 cases with each case being characterized by 24 binarized attributes. These data are referred to as CoIL in the forthcoming.

### 4.3 RESULTS

Figure 1 illustrates the effectiveness of KES for the ALL data when different values for the parameter $k$ are considered. Concretely, it compiles the BIC values, sorted in ascending order, of the BNs resulting from 1000 independent runs of KES for the ALL data when $k = 0, 0.005, 0.02, 0.1, 0.4, 0.8, 1$. Recall that KES with $k = 0$ corresponds to SES and with $k = 1$ to GES. The first conclusion that we can make from Figure 1 (left) is that SES performs poorly. The reason of this somewhat deceptive behavior is in the fact that there exist many locally optimal BNs for the data at hand making, therefore, negligible the probability of SES reaching a model of similar effectiveness to that returned by GES. Specifically, the 1000 independent runs of SES for the ALL data identify 1000 different locally optimal BNs. The second observation that we can make from the graph on the left of Figure 1 is that increasing greediness seems to be beneficial in general. The graph on the right of Figure 1, which is a close-up of the graph on the left, shows that this is true only to some extent: KES with $k = 0.4, 0.8$ outperforms GES significantly often, exactly in 22 % and 66 % of the runs, respectively. This confirms that the trade-off between greediness and randomness enabled by KES can result in a gain of effectiveness when compared to GES, which is based solely on greediness.

Table 1: Effectiveness of KES for the Alarm, Trap, ALL, AML and CoIL databases when $k = 0, 0.4, 0.8, 1$.

| Data | k | Best | >GES | <GES |
|------|---|------|------|------|
| Alarm | 0 | -95207 | 0 | 1000 (1000) |
|  | 0.4 | -94288 | 0 | 918 (477) |
|  | 0.8 | -94288 | 0 | 565 (69) |
|  | 1 | **-94288** | — | — |
| Trap | 0 | **-227932** | 1000 (840) | 0 |
|  | 0.4 | -227955 | 857 (343) | 88 (3) |
|  | 0.8 | -227977 | 275 (39) | 276 (3) |
|  | 1 | -227999 | — | — |
| ALL | 0 | -130154 | 0 | 1000 (1000) |
|  | 0.4 | **-126856** | 222 (222) | 774 (774) |
|  | 0.8 | -126873 | **656** (606) | 324 (316) |
|  | 1 | -127010 | — | — |
| AML | 0 | -74503 | 0 | 1000 (1000) |
|  | 0.4 | -73956 | 32 (32) | 968 (968) |
|  | 0.8 | **-73952** | **227** (131) | 719 (524) |
|  | 1 | -73998 | — | — |
| CoIL | 0 | **-10964** | 11 (11) | 989 (983) |
|  | 0.4 | **-10964** | 114 (43) | 879 (674) |
|  | 0.8 | -10967 | **221** (19) | 740 (126) |
|  | 1 | -10969 | — | — |

Due to space restrictions, the graphs corresponding to the effectiveness of SES, KES ($k \neq 0, 1$) and GES for the rest of the databases in the evaluation are not shown. Instead, Table 1 summarizes the performance of KES ($k = 0, 0.4, 0.8, 1$) for the five databases considered. Unless otherwise stated, KES refers to KES with $k = 0.4, 0.8$ in the discussion below. For each database, Table 1 reports the highest BIC value scored by the BNs induced by each algorithm in the 1000 independent runs, **Best**, and the number of the runs where SES, alternatively KES, performs better, **>GES**, and worse, **<GES**, than GES in terms of BIC values of the induced models. Additionally, **>GES** and **<GES** include in parenthesis the number of distinct models identified in these runs. As can be appreciated from Table 1, KES outperforms GES significantly often for all the databases, except for the Alarm. It is particularly encouraging that KES enjoys such a behavior relative to GES for all the 3 real databases: between 3 % and 22 % of the runs of KES ($k = 0.4$) and between 22 % and 66 % of the runs of KES ($k = 0.8$) identify models that score higher BIC values than that of the model selected by GES. In other words, these results suggest that the probability of KES identifying a locally optimal BN with higher BIC value than that of the BN returned by GES can be considerable, specially for real data. Furthermore, many of the runs where KES beats GES converge to different models. Therefore, it is not only that GES misses the best model in four out of the five databases involved in the evaluation, but that there are several models that are superior to the one returned by GES. These observations indicate that learning BNs based solely on greediness, as GES does, may likely lead to noticeable suboptimal performance in practice. The reason, again, is that many locally optimal BNs may exist, even for data faithful to a DAG.

Going back to the Alarm database, it is not surprising that GES is more effective than SES and KES: the Alarm database is faithful to a DAG and, therefore, GES is asymptotically optimal. The results in Table 1 suggest that the 20000 cases in the Alarm database may be enough to guarantee such a behavior. As a matter of fact, the BN returned by GES is equivalent to the generative model, except for the arc not supported by the data that is missing. Regarding the Trap database, it is not surprising either that SES and KES outperform GES by far. Recall that the Trap database is generated to deceive GES. The fact that SES is superior to KES for this domain stresses that incorporating some kind of randomness in the model learning process may be of crucial importance.

Regarding the number of different locally optimal BNs for the databases in the evaluation, Table 1 provides the reader with evidence that this number is very large. Out of the 1000 independent runs, SES identifies 1000 distinct locally optimal models for the Alarm, ALL and AML databases, 994 for the CoIL and 840 for the Trap. Obviously, these figures are just lower bounds of the number of local optima and, therefore, many more such models may actually exist for these databases. Moreover, even when the data at hand is faithful to a DAG and the amount of available data is considerable, the number of locally optimal BNs may be huge. See the results for the Alarm database in Table 1: faithful to a DAG, large sample assumption supported by the unbeatable behavior of GES, but however at least 1001 different locally optimal models exist.

Finally, it is worth mentioning that running all the experiments in this section with the BDeu (equivalent sample size of 1) as score, instead of the BIC, led to the same conclusions as those discussed above. We also ran the experiments with different sizes of the learning databases. The most interesting observation that we made is that for small sizes of the Alarm database (200-1000 cases for the BIC and 100-500 cases for the BDeu) KES outperformed GES as often as reported in Table 1 for the real databases.

## 5 CONCLUSIONS

We introduced and evaluated a new algorithm, KES, for BN learning from complete data. Actually, KES can be seen as a family of algorithms, due to the fact that it provides the user with the possibility of trading off between greediness and randomness through a parameter $k \in [0, 1]$. We proved that KES, no matter the degree of greediness, i.e. the value of $k$, is asymptotically optimal under the faithfulness assumption as well as asymptotically inclusion optimal when only the composition property holds. When greediness is set at maximum ($k = 1$), KES is equivalent to GES. On the other hand, when greediness is set at minimum ($k = 0$), KES is named SES. We proved that SES can asymptotically discover any inclusion optimal BN with nonzero probability under the composition property assumption. Unfortunately, there can be exponentially many such models for some given data. This limits the applicability of SES. In fact, the behavior of any instance of KES (including GES) can be suboptimal when several inclusion optimal models exist.

The experimental evaluation of SES and KES ($k = 0.4, 0.8$) with respect to GES was carried out by comparing the BIC and BDeu values of the BNs induced from two synthetic and three real databases. The main conclusions made from the results reported are the following ones. First, the number of locally optimal BNs can be huge, even when the faithfulness assumption holds and the amount of available learning data is considerable. Second, SES performs poorly due to the previous point. Third, KES ($k = 0.4, 0.8$) outperforms GES significantly often in general. In particular, it is very noticeable and encouraging that KES ($k = 0.4, 0.8$) beats GES in up to 66 % of the runs for the real databases in the evaluation. Consequently, BN learning algorithms that address the problems derived from the existence of several locally optimal models, for instance, trading off between greediness and randomness as KES does, are worth studying.

### Acknowledgements

We would like to thank the DSS group at Aalborg University for valuable comments to preliminary versions of this paper, Greg Cooper for providing the Alarm database, and Iñaki Inza and Rosa Blanco for making the discretized Leukemia database available.

## References

Beibel M. (2000). Selection of Informative Genes in Gene Expression Based Diagnosis: A Nonparametric Approach, *Proceedings of the First International Symposium in Medical Data Analysis*, pp. 300-307.

Ben-Dor A., Friedman N. and Yakhini Z. (2001). Class Discovery in Gene Expression Data, *Proceedings of the Fifth Annual International Conference on Computational Molecular Biology*, pp. 31-38.

Castelo R. and Kočka T. (2003). On Inclusion-Driven Learning of Bayesian Networks, *Journal of Machine Learning Research*, accepted.

Chickering D. M. (1995). A Transformational Characterization of Equivalent Bayesian Network Structures, *Proceedings of the Eleventh Conference on Uncertainty in Artificial Intelligence*, pp. 87-98.

Chickering D. M. (2002). Optimal Structure Identification with Greedy Search, *Journal of Machine Learning Research*, 3, pp. 507-554.

Chickering D. M. and Meek C. (2002). Finding Optimal Bayesian Networks, *Proceedings of the Eighteenth Conference on Uncertainty in Artificial Intelligence*, pp. 94-102.

Cooper G. and Herskovits E. H. (1992). A Bayesian Method for the Induction of Probabilistic Networks from Data, *Machine Learning*, 9, pp. 309-347.

Cowell R. G., Dawid A. P., Lauritzen S. L. and Spiegelhalter D. J. (1999). *Probabilistic Networks and Expert Systems*, Springer-Verlag, New York.

Golub T. R. and eleven co-authors (1999). Molecular Classification of Cancer: Class Discovery and Class Prediction by Gene Expression Monitoring, *Science*, 286, pp. 531-537.

Herskovits E. H. (1991). *Computer-Based Probabilistic-Network Construction*, PhD Thesis, Stanford University.

Lauritzen S. L. (1996). *Graphical Models*, Clarendon Press, Oxford.

Meek C. (1997). *Graphical Models: Selecting Causal and Statistical Models*, PhD Thesis, Carnegie Mellon University.

Putten P. and Someren M. (eds.). (2000). *CoIL Challenge 2000: The Insurance Company Case*, Sentient Machine Research.

Spirtes P., Glymour C. and Scheines R. (1993). *Causation, Prediction, and Search*, Springer-Verlag, New York.